\documentclass[conference]{IEEEtran}
\usepackage{cite}
\usepackage{amsmath,amssymb,amsfonts}
\usepackage{algorithmic}
\usepackage{graphicx}
\usepackage{textcomp}
\usepackage{xcolor}
\def\BibTeX{{\rm B\kern-.05em{\sc i\kern-.025em b}\kern-.08em
    T\kern-.1667em\lower.7ex\hbox{E}\kern-.125emX}}

\usepackage{pgfplots}
\pgfplotsset{compat=1.17}
\usepackage{pgfplotstable}
\usepackage{svg}
\usetikzlibrary{matrix, positioning}

\begin{document}

\title{Unveiling Interesting Insights: Monte Carlo Tree Search for Knowledge Discovery
}

\author{\IEEEauthorblockN{Pietro Totis}
\IEEEauthorblockA{\textit{J.P. Morgan AI Research} \\
pietro.totis@jpmorgan.com}
\and
\IEEEauthorblockN{Alberto Pozanco}
\IEEEauthorblockA{\textit{J.P. Morgan AI Research} \\
alberto.pozancolancho@jpmorgan.com}
\and
\IEEEauthorblockN{Daniel Borrajo}
\IEEEauthorblockA{\textit{J.P. Morgan AI Research} \\
daniel.borrajo@jpmorgan.com}
}

\maketitle

\begin{abstract}
Organizations are increasingly focused on leveraging data from their processes to gain insights and drive decision-making. However, converting this data into actionable knowledge remains a difficult and time-consuming task. There is often a gap between the volume of data collected and the ability to process and understand it, which automated knowledge discovery aims to fill. 
Automated knowledge discovery involves complex open problems, including effectively navigating data, building models to extract implicit relationships, and considering subjective goals and knowledge. 
In this paper, we introduce a novel method for Automated Insights and Data Exploration (AIDE), that serves as a robust foundation for tackling these challenges through the use of Monte Carlo Tree Search (MCTS). 
We evaluate AIDE using both real-world and synthetic data, demonstrating its effectiveness in identifying data transformations and models that uncover interesting data patterns. Among its strengths, AIDE's MCTS-based framework offers significant extensibility, allowing for future integration of additional pattern extraction strategies and domain knowledge. This makes AIDE a valuable step towards developing a comprehensive solution for automated knowledge discovery.
\end{abstract}

\section{Introduction}
The exponential growth in interest in Artificial Intelligence (AI) across various organizations has highlighted the importance of data collection and analysis. When properly analyzed, data can provide valuable insights into underlying domains and processes. However, extracting these insights is time-consuming and requires a diverse range of expertise, including data manipulation~\cite{data_cleaning} (such as data collection, cleaning, and dataset construction), model building~\cite{model_selection,model_evaluation} (including model selection, evaluation, and parameter optimization), as well as reporting and visualization~\cite{data_visualization}. Automating this process could enable organizations to scale their ability to transform data into knowledge~\cite{automating_data_science}.

Knowledge discovery is challenging because it requires a broad set of analytical skills, involves exploring virtually infinite combinations of data subsets and model-building methods, and includes subjective and domain-specific elements. To automate this process, a system must efficiently navigate through data and models, use the right analytical tools with appropriate confidence levels, differentiate between interesting and trivial findings, and interact with humans for both input and output of knowledge. Each of these aspects presents its own unique challenges, making the automation of knowledge discovery an ongoing effort. Despite significant work from research communities in statistics~\cite{hand1998data}, machine learning~\cite{automl_book}, and data mining~\cite{DBLP:journals/aim/FayyadPS96}, automated knowledge discovery remains an open problem.

In this paper we focus on the problem of efficiently navigating the space of data and models to identify interesting insights.
In Section~\ref{sec:method} we introduce a novel approach for Automated Insights and Data Exploration, AIDE. AIDE balances investigating promising data subsets with exploring new ones, while testing and refining model combinations. Key contributions include:

\begin{itemize} 
\item Framing data exploration and model selection as a \emph{single-player Monte Carlo Tree Search} (MCTS) problem. 
\item Introducing techniques to represent and reason over the infinite action space. 
\item Analyzing the impact of single-player MCTS enhancements from literature. 
\item Proposing domain-specific interestingness measures. 
\end{itemize}

Experiments in Section~\ref{sec:experiments} demonstrate AIDE's effectiveness in identifying meaningful data transformations and model combinations. Section~\ref{sec:related_work} highlights AIDE's novelty compared to previous work. Its modularity allows easy expansion with additional data transformations and discovery techniques. Section~\ref{sec:limitations_future_work} discusses future extensions, including new data and model actions and incorporating domain knowledge to refine interestingness evaluation.

\section{Background}\label{sec:background}
Knowledge discovery is ``the nontrivial extraction of implicit, previously unknown, and potentially useful information from data''~\cite{kdd_foundations}. The information comes in the form of patterns, that is, statements in some language describing relationships among subsets of data. Patterns become \emph{knowledge} when they meet the user's criteria of \emph{interestingness} (non-trivial, novel, useful) and \emph{certainty} (valid with some degree of confidence)~\cite{kdd_foundations}. 
 Knowledge discovery is described in~\cite{DBLP:journals/aim/FayyadPS96} as a multi-step process: (1) understand the application domain and set goals; (2) create a target dataset by selecting relevant data; (3) preprocess the data to handle noise and missing values; (4) reduce and project data to find useful features; (5) match knowledge discovery goals to a suitable data mining method; (6) select models and hypotheses for data analysis; (7) perform data mining to discover patterns; (8) interpret and visualize the mined patterns; (9) act on the discovered knowledge, possibly iterating through previous steps.

Data selection, reduction, and projection can be framed as a search problem to find transformations that improve subsequent  knowledge discovery steps. Similarly, choosing a data mining method and model~\cite{data_mining_book} involves searching for algorithms that best capture data insights. Model evaluation is crucial, especially when steps are automated without human supervision. This motivates our focus on framing data transformations and model choices as a search problem, guided by automatic evaluation of patterns' interestingness.

\subsection{Monte Carlo Tree Search}
Monte Carlo Tree Search (MCTS)~\cite{mcts_survey,mcts_survey_recent} is an algorithm for decision-making problems with a large combinatorial search space represented as a tree, where nodes are problem configurations (states) and edges are actions. MCTS is effective in several domains~\cite{MCTS_app}, in particular games~\cite{mcts_general_game_playing,alphago}, where an exhaustive search of the possible action sequences is infeasible. MCTS balances exploration of new actions and exploitation of promising ones to maximize high-quality decision paths, such as winning a game or minimizing costs. MCTS iteratively builds the search tree through four stages: selection, expansion, simulation, and backpropagation. Selection navigates to a non-terminal leaf node using a tree policy. Expansion adds a new child node from the selected action. Simulation estimates the new node's value with a default policy until reaching a terminal state. Backpropagation updates ancestor nodes' statistics to refine future value estimates.

\textbf{Tree policy.} The tree policy guides the search process selecting the sequence of actions, with the Upper Confidence Bounds applied for Trees (UCT)~\cite{UCT} being a popular approach. In UCT, actions $a$ for a state $s$ are ranked by: $Q(s,a) + C\cdot\sqrt{\frac{\ln{N(s)}}{N(s,a)}}$, where $Q(s,a)$ is the reward estimate, $N(s)$ is the visit count for $s$, and $N(s,a)$ is the count for action $a$ in $s$. The constant $C$, typically $\sqrt{2}$ when $Q$ returns values in $[0,1]$~\cite{mcts_survey_recent}, balances exploitation of promising nodes (high $Q(s,a)$) and exploration of less tested actions (low $N(s,a)$).

\textbf{Single-player MCTS.} MCTS in single-player games~\cite{MCTS_single}, differs from two-player games due to a wider outcome range and no opponent uncertainty. Extreme reward estimates are risky in two-player MCTS because they may come from overly optimistic assumptions about the opponent strategy, but this is not a factor in single-player MCTS. Accounting for the variance of the child estimate in the UCT policy~\cite{MCTS_single}, increases scores for actions leading to extreme reward estimates. The authors in~\cite{mcts_survey} formulate this as adding $\sqrt{\sigma^2+\frac{D}{N(s,a)}}$ to the standard UCT policy, where $\sigma^2$ denotes the child's estimates variance, and $\frac{D}{N(s,a)}$ serves to inflate the term for less visited children, with a fixed constant $D$ similar to $C$.

\textbf{Transpositions.} In domains like chess, different action sequences can lead to the same state, forming a directed acyclic graph. The authors in~\cite{transpositions} propose adaptations to this setting for UCT, to account for the visits and consequent rewards that a child may receive from other parents. For instance, UCT2 is: $Q(g(s,a)) + C\cdot\sqrt{\frac{\ln{N(s)}}{N(s,a)}}$, using $Q(g(s,a))$, the average reward for all estimates of $a$ in the child state, instead of just those from action $a$ in $s$.

\textbf{Progressive widening.}
In domains with large action sets, exploring every possible action is impractical. UCT-based policies, however, perform well when they can refine the reward estimates of each action multiple times. \emph{Progressive widening}~\cite{mcts_pw2,mcts_pw} addresses this challenge by initially exploring only a subset of actions and gradually increasing the subset size based on the node's visit count, governed by the parameter $\alpha\in(0,1)$: when $\lfloor N(s)^\alpha \rfloor \geq |\mathit{children}(s)|$ a new action is activated. Strategies for selecting new actions to explore include using scoring functions~\cite{mcts_fpu} or sampling from continuous distributions~\cite{double_pw}.

\subsection{Measuring interestingness}
One of the main challenges of extracting interesting insights from data is defining and measuring interestingness. A recent survey of the literature~\cite{cube_interesting}, provides a taxonomy and analysis of interestingness in data cube queries, concluding that interestingness is a combination of multiple dimensions: relevance, surprise, novelty, and peculiarity. Relevance relates to the user's information goal, surprise assesses deviation from user expectations, novelty indicates information unknown to the user, and peculiarity measures how different an object is from its peers, akin to outliers. Peculiarity is unique in being the only dimension measurable without user input.

\section{Method}\label{sec:method}

\begin{table}[b]
\caption{Tabular dataset $D_0$ example}
\centering
\begin{tabular}{|c|c|c|c|}
\hline
\textbf{Student} & \textbf{Exam Date} & \textbf{Score} & \textbf{Course Duration} \\
\hline
S1 & 2020-09-01 & 88          & 1 year           \\\hline
S2 & 2019-08-15 & 92          & 3 months          \\\hline
S3 & 2021-01-10 & 75          & 6 months         \\\hline
S4 & 2018-05-20 & 85          & 1 year       \\\hline
\end{tabular}
\label{tab:example}
\end{table}

We propose AIDE, a method to tackle some of the key challenges in automating knowledge discovery. It focuses on (1) efficiently identifying data transformations to improve pattern mining (step 4 of the knowledge discovery process), (2) selecting methods that best capture implicit knowledge (steps 5-7), and (3) iterative refinements (steps 8-9). The first challenge involves efficiently evaluating infinite data transformation options. The second challenge deals with the complexity and high computational cost of pattern mining, requiring a targeted application of data mining techniques. The third challenge involves the subjective
nature of measuring interestingness, requiring a reliable unsupervised estimate.

Let $D_0=(C_1,\dots, C_n)$ be a tabular dataset with $n$ columns $C_i\in D_0$. We also denote a value $v$ belonging to a column $C_i$ as $v\in C_i$. 
We assume the data is clean and preprocessed, although our implementation can handle null values. This includes the assumption that each column $C_i$ is associated to one of the following data types: numerical, datetime or timedelta (quantitative types), and categorical or boolean (qualitative types). 
For example Table~\ref{tab:example} is a tabular dataset with columns Student (categorical), Exam date (datetime), Score (numerical) and Course Duration (timedelta).
We will define transformations of columns that generate new columns from the original ones, denoting with $D_0^*$ the closure of the set of original columns in $D_0$ with respect to the given transformations. The goal of AIDE is to automatically discover interesting relationships within $D_0^*$.  We express as $\mathcal{D}$ the set of all possible subsets of columns, rows, or both, of $D_0^*$. 
To extract relationships from $D_0^*$, we consider the combinations of subsets $D\in\mathcal{D}$ with a set $\mathcal{M}$ of data mining models or techniques, such as decision trees, k-means clustering, outlier detection, etc. We add to both $\mathcal{D}$ and $\mathcal{M}$ the symbol $\varnothing$ to denote respectively an empty dataset and no model. Section~\ref{sec:states} defines the space $\mathcal{D}\times\mathcal{M}$. 
A model $M\in\mathcal{M}$ fitted on $D\in\mathcal{D}$, denoted by $M_D$, corresponds to a pattern $P_M^D$, which we consider to be a textual summary of the relationships encoded in $M_D$.
Section~\ref{sec:search} discusses our pattern search strategy in $\mathcal{D}\times\mathcal{M}$.
While defining functions $str: \mathcal{D}\times\mathcal{M}\rightarrow P_M^D$ that map fitted models to text is important for user interaction, it is not the primary focus of this work. 
However, we address this aspect in our implementation with pre-defined textual templates.
Finally, let $\mathcal{P}$ be the set of all patterns. The interestingness of $P_M^D\in\mathcal{P}$ is $intr(P_M^D)$, where $intr$ is a function $intr: \mathcal{P}\rightarrow[0,1]$. Section~\ref{sec:interesting} focuses on the definition of $intr$.

\subsection{States and actions}\label{sec:states}

\textbf{States.} We represent a state as a pair $(D,M)\in \mathcal{D}\times\mathcal{M}$.
This pair represents the combination of the exploration of data transformations with the application of pattern extraction models.
A dataset is determined by a sequence of data transformations, but different sequences can lead to the same dataset. When equivalent sequences for a dataset $D$ are paired to the same model $M$, we have a transposition in $(D,M)$. A state change occurs when either the dataset is further transformed or a different model is fitted. In fact, we distinguish two families of actions: the data actions and the model actions. 

\textbf{Data actions.} Data actions describe transitions $(D,M)\rightarrow(D',\varnothing)$. When the search moves to a different data subset, the associated model is $\varnothing$ because no model has been fitted to $D'$ yet. We define four types of data actions that transform the dataset: \emph{select}, \emph{derive}, \emph{where}, and \emph{groupby} actions. 
\emph{Select} actions join to the current dataset $D\in\mathcal{D}$ one of the original columns in $D_0$. For example \emph{select}(Student) if Student is not in $D$. 
\emph{Derive} actions generate a new column by transforming one or more columns in the current dataset $D$. The derivation primitives are unary or binary operations. Unary operations are value discretization operations, that is, discretizing a column by equal-size bins or quantiles (on numerical and timedelta columns), or discretizing a datetime column by one of its attributes, i.e. year, month, day, hour, etc. For example $\mathit{derive}(\text{Score}, \mathit{bins}, 10)$ generates a categorical column where each row is assigned to one of 10 equally-sized bins based on the Score value.   $\mathit{derive}(\text{Exam Date}, \mathit{time}, \mathit{month})$ generates a column containing the month of each Exam Date. 
Binary operations are element-wise operations of the kind $C_i\ast C_j$. They include arithmetic operations, i.e. $\ast\in\{+,-,\cdot,\div\}$ and comparison operations, i.e. $\ast\in\{=,\neq,>,<\}$. For example, a possible binary arithmetic operation in Table~\ref{tab:example} is: $\text{Exam Date} - \text{Course Duration}$. \emph{Where} actions define filtering operations on the dataset. A \emph{where} action is of the kind $C \ast v$, where $C$ is a column, $\ast\in\{=,\neq,>,<\}$ and $v\in C$ and filters the rows that satisfy the condition. For instance, $\mathit{\mathit{Score} < 60}$ generates a subset of Table~\ref{tab:example} with all the students' exams with score less than 60. Finally, the \emph{groupby} action is of the kind $\mathit{group}(G,agg_1(C_1),\dots,agg_n(C_n))$ where $G\in D$ is the column defining the groups, $C_1,\dots,C_n$ are the remaining columns to be aggregated into a single value for each group, and $agg_1,\dots,agg_n$ are the corresponding aggregation functions. 
The options for $agg_i$ include: $\mathit{min, max, avg, median, sum, std}$ on quantitative columns, $\mathit{all}$, $\mathit{any}$ for boolean columns, and $\mathit{mode}, \mathit{freq}(C_i,v)$ for categorical columns. $\mathit{freq}(C_i,v)$ is a function counting the frequency of value $v\in C_i$ of each group. For instance, in Table~\ref{tab:example} $
\mathit{group}(\text{Student},\allowbreak \mathit{max}(\text{Exam Date}),\allowbreak \mathit{avg}(\text{Score}),\allowbreak \mathit{freq}(\text{Course Duration},\allowbreak \text{6 months}))$
generates a new data subset by aggregating each student's data with the last exam date, the average exam score, and the number of 6 months-long courses taken.
Note that a \emph{groupby} action transforms the columns, meaning that the corresponding subtree selects columns of the aggregated dataset rather than $D_0$. Moreover, sequences of data actions are monotonic on any tree branch: once a column is added to $D$ to any state, it cannot be removed from any posterior state in the same branch. And once a row is filtered out of $D$ with a \emph{where} action, it cannot be reintroduced.

\textbf{Model actions.} Model actions describe transitions $(D,M)\rightarrow(D',M')$. A model action can be any data mining technique. We implemented some of the most common, such as: computing association rules, fitting a decision or regression tree for target columns, analysing a time series for datetime columns and target features, finding anomalous values (outliers) and clustering. While for most model actions $D=D'$, outlier detection and clustering modify the dataset in the state transition by adding a column, i.e. $D\neq D'$. The former adds a boolean column with labels indicating if the entry is an outlier w.r.t. the given reference feature(s). The latter adds a numerical column with the cluster identifier assigned to each entry. The choice of target columns for trees or time series analysis, as well as other hyperparameters of the model actions, e.g. tree depth or number of clusters, are part of the search process (Section~\ref{sec:search}).

\textbf{Action representation.} Representing and reasoning over data and model actions is not trivial due to the large number of parameter combinations, and the corresponding applicability rules. To avoid an inefficient enumeration and evaluation of all possible actions at each node, we represent actions as labelled trees, where nodes denote parameters and edges represent values. Nodes are either ground, if labelled with a single value, or lifted, if labelled with a set of values. Figure~\ref{fig:actions_small} represents the (lifted) nodes for each action type and the corresponding possible ground values. 
Actions are root-leaf paths where each parameter corresponds to a ground value.
The action instantiation logic is defined by how the next child is selected, and, when the child is lifted, how the ground label value is chosen. 
Lifted nodes implicitly represent exponentially large combinations of parameter choices allowing a more efficient precondition evaluation. 
Some preconditions determine the feasibility of the action, such as preventing the summation of two columns with incompatible types (hard preconditions). Others avoid the creation of low-quality states, such as blocking \emph{where} actions that would filter out all data (qualitative preconditions). Additionally, some preconditions prevent the repetition of similar actions, like blocking the commutative counterpart of binary \emph{derive} actions (search preconditions).

\begin{figure}[t]
\centering
\small
\begin{enumerate}
    \item \textbf{Select parameter:} Column
    \item \textbf{Derive}
    \begin{enumerate}
        \item \textbf{Discretize parameters:}
        \begin{itemize}
            \item \textbf{target:} Column
            \begin{itemize}
                \item \textbf{bins:} $\{2,\dots,10\}$
                \item \textbf{quantiles:} $\{2,3,4,5,10\}$
                \item \textbf{time:} $\{\mathit{year}, \mathit{quarter}, \dots, \mathit{minute}, \mathit{seconds}\}$
            \end{itemize}
        \end{itemize}
        \item \textbf{Binary Operation parameters:}
        \begin{itemize}
            \item \textbf{operator:} $\{+,-,\cdot,\div,=,\neq,>,<\}$
            \begin{itemize}
                \item \textbf{operands:} Left Column, Right Column
            \end{itemize}
        \end{itemize}
    \end{enumerate}
    \item \textbf{Where parameters:} Column, $\{=,\neq,>,<\}$, Value
    \item \textbf{Groupby parameters:}
    \begin{itemize}
        \item \textbf{grouper:} Column
        \item \textbf{for each column:} $\{\mathit{min, max, \dots, any, mode, freq}\}$
        \begin{itemize}
            \item \textbf{if freq function:} Value
        \end{itemize}
    \end{itemize}
    \item \textbf{Decision Tree parameters:}
    \begin{itemize}
        \item \textbf{target:} Column
        \begin{itemize}
            \item \textbf{max depth:} $\{2,3\}$
        \end{itemize}
    \end{itemize}
    \item \textbf{Unary Outliers parameter:} Column
    \item \textbf{Binary Outliers parameter:} Column Pair
    \item \textbf{Clustering parameters:}
    \begin{itemize}
        \item \textbf{clusters number:} $\{2,\dots,10\}$
    \end{itemize}
    \item \textbf{Trend parameters:} Datetime Column, Target Column
    \item \textbf{Association Rules parameters:} none
\end{enumerate}
\caption{Parameter order and indentation are levels in the action tree.}
\label{fig:actions_small}
\end{figure}

\subsection{Search}\label{sec:search} We approach the problem using a single-player Monte Carlo Tree Search (MCTS) due to its proven ability to effectively balance exploration and exploitation in large combinatorial search spaces, without requiring any prior domain knowledge. Our search process begins with an initial state $(\varnothing,\varnothing)$, representing an empty dataset without any associated model. 
As the search progresses, the dataset is expanded through data action transitions. For instance, in the first iteration of the algorithm, only a \emph{select} action is applicable, which begins introducing columns from $D_0$ into the states. 
This approach enables the knowledge discovery process to start by searching for simpler patterns using a reduced set of features, and then gradually progress to more complex patterns involving a larger number of features. By limiting the scope of features in the data mining models, we can initially explore many small models with less computational effort than if we used the full set of features. At the same time, this strategy allows the MCTS to direct more computational resources later towards larger sets of features that expand early promising results. The search discovers a successful action path when it reaches a state corresponding to a pattern with interestingness above a predefined threshold, such as 0.5. 
The algorithm concludes the search upon reaching the specified number of MCTS iterations. This is the only termination condition in this domain, as only an exhaustive search can ensure that all potential patterns, for the given set of actions, have been extracted. Unlike games or optimization problems, we are not interested in a single winning or optimal decision path; instead, the goal is to collect as many successful paths as possible. The MCTS stages are as follows.

\textbf{Selection.} We implement different tree policies: random, UCT, UCT for single player-games, and the variant for transpositions. For each state $(D,M)$ we keep track of the actions with valid preconditions, to avoid selecting terminal states where no actions are applicable. Generating a terminal state becomes extremely unlikely when the algorithm avoids the actions filtering out all the data, because it would mean that no new data transformations are possible and that all available models have been tested. 

\textbf{Expansion.} Despite filtering actions with preconditions, the available actions at each expansion step are virtually infinite, making it infeasible to test all valid actions. Therefore, we have to select a set of actions actively searched. We consider two strategies to determine the number of actions actively searched at each node: using a constant number of active actions, and progressive widening. In both cases, the question is how to select the next action to add to the actively searched set.

A basic strategy is to randomly select the action. In this domain this approach is not particularly penalizing because an action inactive at a node $(D,M)$ is often applicable later in its subtree. For example, selecting $C_1$ over $C_2$ does not prevent testing $\mathit{select}(C_2)$ later in the $\mathit{select}(C_1)$ subtree. The exception is the \emph{groupby} action: grouping by $C_1$ rather than $C_2$ means that grouping by $C_2$ is no longer possible in the subtree for $\mathit{group}(C_1,\dots)$. 
However, we aim to select effective actions early, to increase the likelihood of encountering states with interesting patterns within the iteration limit. 

In standard MCTS, a leaf node lacks information beyond the actions on the path to the initial state. The key idea to improve random selection is to leverage information about actions in other subtrees to estimate the effectiveness of the available actions. In many cases, the interchangeability of action order encourages sharing information about promising actions across different subtrees. For example, if a node expands $\mathit{where}(C_1<v)$ and the resulting child is interesting, other nodes should be encouraged in trying the same action or exploring some of its parameters, for instance by selecting $C_1$ or filtering a different column relative to $v$. This strategy also has a practical motivation, as data scientists are often biased towards features or parameters that appear to be frequently associated with interesting results, or come from past experience.

We assume that the type of action alone does not inherently determine the interestingness of the outcome. Therefore, we begin with a random selection of the action type. Following this, we define the action by selecting the corresponding parameters. To account for the potential influence of these parameters on interestingness, we weight the random selection of each parameter value based on its expected contribution to interestingness. This contribution is estimated by averaging the difference in interestingness between a child node and its parent node when the connecting action involves the parameter value. 
While this approach might miss the complex interactions of the whole action sequence and the different importance of each parameter, it offers a practical way to improve over random selection, without trying to solve a credit assignment problem~\cite{credit_assignment} at each expansion.
A further improvement to balance exploitation of promising parameter values with exploration of untested values, is to add a visit count to the values and replace the weighted random choice with a UCT-based choice.

\textbf{Simulation.}
The simulation step estimates a node's expected reward using a default strategy, providing an initial interestingness estimate. We use a simulation strategy less effective than data mining models, but it requires less computation. In this domain and single-player context, it is not necessary to reach a terminal state and there is no reason to discard the simulation result. 
We assign a newly created node $(D,\varnothing)$ a base score based on unary and binary statistics of $D$. Unary statistics are standard statistical summaries: for quantitative columns we compute entropy, skeweness, kurtosis and the interquartile range to mean ratio, while for qualitative types we compute entropy, number of unique values and the most common value with its frequency. Similarly, binary statistics measure the pairwise correlation of the columns, rewarding datasets $D$ with a large number of highly correlated pairs of columns. 
We aggregate these measures to derive an interestingness score for the new node (more details in Section~\ref{sec:interesting}).

\textbf{Backpropagation.}
The backpropagation step is a standard recursive update of the parent score based on the new children estimates. To aggregate the children score we consider the mean, the typical strategy in MCTS, and the root mean square (\emph{rms}). We experiment with \emph{rms} because this function amplifies the impact of larger values on the aggregated score. This effect reflects the importance of a very promising child, even among many others lowering the overall mean with low interestingness estimates. We also update the score of each action with the difference in the new parent-child estimates, to refine the next parameter selection of new active actions.  

This method tackles challenges 1 and 2 by using MCTS to explore data transformations and select suitable data mining techniques, allowing the MCTS to focus on promising data subsets and limit model scope, reducing computational effort. We now address the third challenge, estimating interestingness, crucial for MCTS effectiveness.

\subsection{Interestingness}\label{sec:interesting}
In an unsupervised domain, interestingness estimation relies on the dimension of peculiarity, because the other dimensions (relevance, surprise, and novelty) depend on contextual information and user knowledge, which we assume unavailable in this setting. Therefore, in this section we define interestingness based on different forms of peculiar patterns. We define $intr: \mathcal{P}\rightarrow[0,1]$ differently for each pattern detection method $M$, as the notion of peculiarity depends on the type of data relations detected. We require all estimates to be in $[0,1]$ to allow the MCTS to compare and aggregate values from different action types.

\textbf{Simulation.} The simulation in a state $(D,\varnothing)$ assesses interestingness with univariate and bivariate statistics. We estimate the interestingness of a single column based on how ``peculiar'' the value distribution is. For quantitative columns, we use skewness to detect asymmetries, kurtosis to identify long tails and potential outliers, and the ratio of the interquartile range to the mean to highlight distributions with unusual spread relative to the average. For qualitative columns, we consider a distribution peculiar if the most common value has a frequency greater than $90\%$, indicating uncommon categories, or if there is an even distribution of categories, which we assess with their entropy. In terms of bivariate analysis, we start with the assumption that there are no inherent relationships in the data, so any strong correlation between two columns is peculiar. The binary statistics component of the simulation's interestingness estimate is the proportion of column pairs with a Pearson correlation coefficient greater than $0.8$.
Once a new node is simulated and assigned a first interestigness measure, the MCTS decides if and when to apply a model action to refine this estimate. We now define $intr(P_M^D)$ for a state $(D,M)$ with pattern $P_M^D$, depending on $M$.

\textbf{Trees.} If $M_D$ is a decision or regression tree for a target column $C\in D$, interestingness is the product of three metrics: $intr(P^D_M)=acc(M_D)\cdot ent(C)\cdot cover(M_D,C)$. $acc(M_D)$ is the $f_1 \mathit{macro}$ score for decision trees and the $R^2$ score for regression trees. However, trees are prone to overfitting. Therefore, the normalized entropy~\cite{norm_entropy} of the target class $ent(C)$ lowers the interestingness when the target class is highly unbalanced. Because we limit the tree depth to focus on simple relations, a pruned decision tree may not include all target classes. Therefore, $cover(M_D,C)$ measures the proportion of values of $C$ that $M_D$ is capable of predicting, lowering the score for not predicting some categories. For regression trees $cover(M_D,C)=1$.

\textbf{Univariate outliers.} When $M_D$ is univariate outliers detection, we distinguish again between quantitative and qualitative columns. For a qualitative column $C\in D$, we consider any category that is not the most frequent as an outlier if the most frequent category has a frequency greater than $t_{qual} = 0.85$. Being low frequency is not sufficient because in a column of unique identifiers all values would be considered outliers. For quantitative columns, we compute the standard scores (z-scores) relative to the sample mean and standard deviation and consider outliers all values with scores higher than $t_{quant} = 2$. Let $\mathit{out}(v)$ denote such scores for any value $v\in C$. From these scores we synthesize a fine-grained interestingness measure, comparable across different features. For an outlier value $v$, we define this measure as $\mathit{outlier}(v) = 1-\frac{\mathit{out}(v)}{1-t_{qual}}$ if $v$ is qualitative, and $\mathit{outlier}(v) = 1-\frac{t_{quant}}{out(v)}$ if $v$ is quantitative. The first measure is closer to $1$ when the frequency of $v$ gets closer to $0$. In the quantitative case, the measure approaches 1 as the deviation of $v$ from the mean ($out(v)$) becomes increasingly large.
In both cases if $v$ is not an outlier, $\mathit{outlier}(v) = 0$. We define $intr$ as $intr(P_M^D)=0$ if there are no outliers, otherwise $intr(P_M^D)=0.5+\frac{\max_{v\in C} \mathit{outlier(v)}}{2}$ in $C$. This assigns a baseline interestingness of $0.5$ to columns with any type of outliers, which is increased with a peculiarity measure of the most anomalous value.

\textbf{Bivariate outliers.} When $M_D$ is bivariate outlier detection, we focus on feature pairs $(C_i,C_j), i\neq j$ with high correlation (computied during the simulation) to identify value pairs $(v_i,v_j)$, $v_i\in C_i, v_j\in C_j$, that deviate from this correlation. Each value pair is scored based on their correlation, and we isolate the anomalous pairs with  the same approach of the numerical univariate outlier detection. The score is defined differently for pairs of qualitative features, and for mixed or quantitative pairs. For the former we identify under or over-correlated categories pairs by constructing a contingency table and scoring each row and column with the respective Gini index, $\mathit{gini(v_i,v_j)}$. Categories with an anomalous Gini index are under or over-associated with categories of the other feature.  For mixed or quantitative pairs, features are encoded numerically, and value pairs are scored by fitting a linear regression model and computing residuals, $\mathit{residual}(v_i,v_j)$. Residuals outliers suggest anomalous pairs among the correlation. Interestingness is thus: $intr(P_M^D)= \max_{(v_i,v_j), v_i\in C_i, v_j\in C_j} \mathit{corr\_score}(v_i,v_j)$ where $C_i,C_j \in D$ have a correlation coefficient higher than 0.8, and $\mathit{corr\_score}$ is $\mathit{gini}$ if $C_i$ and $C_j$ are categorical, $\mathit{residual}$ otherwise.

\textbf{Clustering.} Clusters' interestingness is based on their separation degree. We use k-means clustering and its silhouette score ($\mathit{silhou}$)~\cite{data_mining_book} normalized in $[0,1]$. However, k-means is sensitive to (encoded) categorical values, therefore we reduce the clustering interestingness depending on the degree of correlation with categorical columns (denoted by $cat(D)$): $intr(P_M^D)=\frac{1+\mathit{silhou}(M_D)}{2}\cdot \max_{C\in cat(D)} corr(C,C_{cluster})$, where $C_{cluster}$ is the column containing the cluster assignment and $corr$ the correlation measure.

\textbf{Trends.} When $M_D$ is a time series analysis, we evaluate if the trend is non-stationary~\cite{pymankendall} ($\mathit{trend}(M_D)=1$), it is periodic~\cite{statsmodel} ($\mathit{period}(M_D)=1$), or it presents outliers~\cite{ts_outliers} ($\mathit{outliers}(M_D)=1$). If a property is not detected, the corresponding function returns $0$. If none of them is detected then $intr(P_M^D)=0$. Otherwise, we assign an interestingness baseline of $0.5$ : $intr(P_M^D)=0.5 + \frac{\mathit{trend}(M_D)+\mathit{period}(M_D)+\mathit{outliers}(M_D)}{6}$.

\textbf{Association rules.} When $M_D$ is an association rules mining model, we use a Kulczynski measure together with the imbalance ratio to evaluate the mined rules~\cite{data_mining_book}. Let $A$ and $B$ be two itemsets for which $M_D$ detected a rule $A\Rightarrow B$: we denote with $kulc(A,B)$ its Kulczynski measure and $ir(A,B)$ its imbalance ratio. A rule is not interesting when $kulc(A,B)=0.5$ or $ir(A,B)=0$. Therefore, we define $intr$ as $intr(P_M^D) = max_{A\Rightarrow B \in M_D} kulc(A,B)\cdot (1-ir(A,B))$.

\section{Experiments}\label{sec:experiments}
The experiments aim to evaluate the effectiveness of the proposed method at discovering interesting data transformations and pattern mining models. We articulate this goal on three questions: (Q1) Can AIDE recognize interesting relations within the data? (Q2) How efficient is it in doing so? (Q3) How do the different MCTS techniques influence these results? To answer these questions we generated different datasets of varying size, including in each synthetic dataset a set of pre-defined patterns, and we evaluated if and how AIDE recognizes these patterns. To have a qualitative type of evaluation, we also analysed which insights AIDE extracts from a set of well-known datasets in the UCI Machine Learning Repository~\cite{uci_machine_learning_repository}.

\textbf{UCI Machine Learning Repository datasets.}
This approach complements our quantitative analysis with synthetic data by providing a familiar context for assessing the algorithm's insights. On Iris~\cite{iris_53} AIDE identifies the correlations of petal length with petal width and sepal length, along with petal length being a good predictor for the class.  On Adult~\cite{adult_2} AIDE detects well-separated clusters between `capital-gain', `capital-loss', `education-num', which relates to the known correlation between education and income level. On Wine quality~\cite{wine_quality_186} AIDE identified a moderate correlation between high `residual\_sugar' and `density' and between high `volatile\_acidity' and `color=red'. On Hearth Disease~\cite{heart_disease_45} AIDE found that normal ECGs often align with no heart disease, specific chest pains link to ECG changes, and higher exercise capacity ties to normal ECGs. 
On Diabetes~\cite{diabetes_130-us_hospitals_for_years_1999-2008_296} AIDE discovered moderate correlations suggesting that patients without a medical speciality often have many diagnoses, or that emergency admissions rarely lead to readmissions, highlighting key healthcare patterns.
In general, AIDE effectively revealed both weak and strong patterns across these datasets that align with known domain insights.

For a quantitative analysis, we tested AIDE's performance in two scenarios. Scenario 1 adds a single pattern to datasets with increasing columns to evaluate scaling in finding one pattern. Scenario 2 adds multiple patterns to datasets with increasing columns and rows to assess scaling in finding multiple patterns. Despite controlling data generation, creating datasets with clear ground-truth and increasing difficulty is challenging. Ensuring only pre-defined patterns exist is difficult, as filtered rows or derived columns may reveal unexpected properties, and different correlation rules on the same feature may correlate generated columns. Modulating search space complexity is also challenging; larger datasets expand the search space, but more patterns increase discovery chances. Thus, designing an experimental setup with a clear metric is complex. However, expected patterns provide a reference for AIDE's discoveries.

\begin{table}[b]
    \caption{Synthetic data: column counts except `Rows' and `Cluster $n$'.}\label{tab:synth_data}
    \centering
    \begin{tabular}{|c|c|c|c|c|c|}
    \hline
        & $\mathcal{SD}_A$ & $\mathcal{SD}_B$ & $\mathcal{SD}_C$ & $\mathcal{SD}_D$ & $\mathcal{SD}_E$ \\\hline
        Columns & 5 & 10 & 15 & 20 & 30 \\\hline
        Rows & 1000 & 5000 & 10000 & 20000 & 30000 \\\hline
        Random & 1 & 4 & 8 & 10 & 14 \\\hline
        Correlation & 1 & 2 & 1 & 2 & 4 \\\hline
        Outliers  & 1 & 2 & 2 & 3 & 4 \\\hline
        Cluster & 1 & 1 & 1 & 2 & 2 \\\hline
        Cluster $n$ & 2 & 3 & 5 & 5 & 6 \\\hline
        Trend & 0 & 1 & 1 & 1 & 2 \\\hline
        Rules  & 1 & 0 & 1 & 2 & 4 \\\hline
        Partial Rules  & 0 & 1 & 2 & 2 & 3 \\\hline
    \end{tabular}
\end{table}

\textbf{Scenario 1.} We created 10 datasets $\mathcal{SD}_i, i\in\{1,\dots,10\}$, each with 1000 rows, starting with two features: a skewed datetime series and a correlated numerical feature with an increasing trend, but containing outliers. We added to $\mathcal{SD}_i$  $(i+1) \cdot 2$ normally distributed columns. We tested different configurations of AIDE (Table~\ref{tab:configs}), computing the minimum number of iterations to recognize the insight, averaged over 10 different random seeds and limited to 500 iterations. The insight can be found with a time-series analysis revealing both the trend and the outliers, or an outlier analysis combined with association rules or decision trees showing column correlation. In this experiment speed depends on the early \emph{select} actions close to the root: selecting the two columns early increases noticeably the discovery chances. The initial actions at the root are random for all configurations because the MCTS has little to no information. Therefore, this experiment highlights which configurations can recover faster from poor initial choices. The results show high variability due to the strong dependency on the initial randomness, with close results across most configurations, therefore we computed across all datasets the average ranking of each configuration (Table~\ref{tab:ranking_scenario1}). The experiment shows progressive widening is a key factor, with configurations C8 and C9 performing worst together with the random policy C1. C1 is also the one showing less variability in being among the worst performing configurations.

\textbf{Scenario 2.} We created five datasets (Table~\ref{tab:synth_data}) starting with random features of different types and adding modifications. We included correlated columns with linear, logarithmic, or exponential relationships, varying in strength and noise. We added outlier features, with numerical values showing outliers and categorical values having a dominant category. We incorporated features clearly separable into distinct groups (cluster columns), and trend columns with increasing, decreasing, or periodic trends relative to a reference datetime. Finally, we defined rule columns with values based on specific rules, such as $\mathit{feature}_1 > 0.7 \rightarrow \text{Category\_A}$, and included partial rules to modify random columns.

\begin{table}[t]
\centering
\caption{Experimental Configuration. \protect\textnormal{Note: R = Random, WR = Weighted Random, RS = Random Simulation, PW = Progressive Widening}}\label{tab:configs}
\begin{tabular}{|c|c|c|c|c|c|c|}
\hline
\textbf{Config.} & \textbf{MCTS} & \textbf{Action} & \textbf{RS} & \textbf{PW} & \textbf{Growth} & \textbf{Fan} \\
& \textbf{Policy} & \textbf{Policy} & &  & \textbf{Rate} & \textbf{Out} \\
\hline
C1 & R & R & Yes & Yes & 0.5 & / \\
\hline
C2 & UCT & R & Yes & Yes & 0.5 & / \\
\hline
C3 & UCT & R & No & Yes & 0.5 & / \\
\hline
C4 & UCT & WR & No & Yes & 0.5 & / \\
\hline
C5 & UCT & UCT & No & Yes & 0.25 & / \\
\hline
C6 & UCT & UCT & No & Yes & 0.5 & / \\
\hline
C7 & UCT & UCT & No & Yes & 0.75 & / \\
\hline
C8 & UCT & UCT & No & No & / & 3 \\
\hline
C9 & UCT & UCT & No & No & / & 6 \\
\hline
C10 & spUCT & UCT & No & Yes & 0.5 & / \\
\hline
\end{tabular}
\label{tab:policy_parameters}
\end{table}

\begin{table}[b]
    \centering
    \caption{Average rank by dataset (lower is better)}
    \label{tab:ranking_scenario1}
    \begin{tabular}{|c|c|c|c|c|c|c|c|c|c|c|}
    \hline
         Conf. & C4 & C3 & C6 & C10 & C7 & C2 & C5 & C9 & C1 & C8 \\ \hline
         Avg. & 2.9 & 3.8 & 4.1 & 4.5 & 4.9 & 5.8 & 5.9& 7.3 & 7.3 & 8.5 \\\hline
         $\pm$ & 2.4 & 1.9 & 2.4 & 3.1 & 2.6 & 2.7 & 2.8 & 2.3 & 1.7 & 2.4 \\ 
         \hline
    \end{tabular}
\end{table}

\begin{figure*}[t]
    \centering
    \includesvg[width=0.98\textwidth]{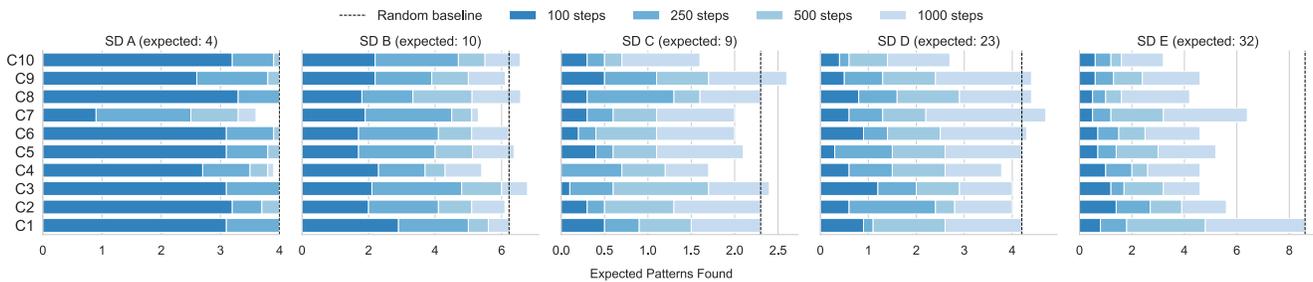} 
    \vspace{-10pt}
    \caption{Expected number of expected patterns discovered by configuration and number of steps.}
    \label{fig:found_plot}
\end{figure*}
\begin{figure*}[t]
    \centering
    \includesvg[width=0.98\textwidth]{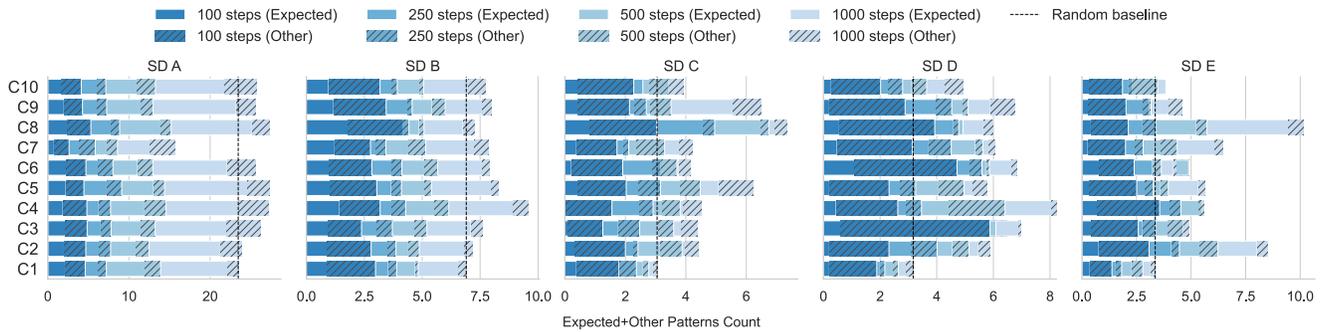}
    \vspace{-10pt}
    \caption{Expected number of nodes reporting expected patterns and other interesting patterns by configuration and number of steps.}
    \label{fig:count_plot}
\end{figure*}

We initially focused on how many expected patterns different configurations of AIDE (Table~\ref{tab:configs}) could find, expecting UCT-based methods (C2-C9) and single-player UCT (C10) would outperform the random node choice policy (C1) given the same MCTS iterations. Let $Exp$ be the set of expected patterns and $Other$ the set of interesting but unexpected patterns. We used four cut-off steps $s\in{100, 250, 500, 1000}$, testing each dataset $\mathcal{SD}_i$, step $s$, and configuration $C_j$ with 10 random seeds $r$, averaging results over 10 runs $\rho = (\mathcal{SD}_i,s,C_j,r)$. For each pattern $p$, we calculated $\mathit{count^{\rho}(p)}$ as the number of nodes reporting $p$ and $\mathit{found}^{\rho}(p)= \min(\mathit{count}^{\rho}(p), 1)$. The expectation of finding $p$ for $SD_i$ with $C_j$ within $s$ iterations is the mean $E_{C_j}^s(\mathit{found}^{\rho}(p))=\sum_{r\in\{1,\dots,10\}}\frac{\mathit{found}^{\rho(r)}(p)}{10}$.

Figure~\ref{fig:found_plot} shows the progression of each configuration $C_j$ over steps $s$ with expected patterns found, stacking $\sum_{p\in \mathit{Exp}} E_{C_j}^s(\mathit{found}^{\rho}(p))$. Contrary to expectations, the random policy $C_1$ often outperforms other configurations, notably in $\mathcal{SD}_E$. We investigated further these results and found an explanation by looking at $\mathit{count^{\rho}(p)}$ and $\mathit{Other}$.

Figure~\ref{fig:count_plot} illustrates the progression of pattern counts, distinguishing expected (non-hatched) from others (hatched), stacking $\sum_{p\in \mathit{Exp}\cup\mathit{Other}} E_{C_j}^s(\mathit{count}^{\rho}(p))$. UCT MCTS configurations consistently find more interesting patterns, both expected and unexpected, especially in the early iterations. Many patterns are association rules involving feature pairs with expected relations, indicating the algorithm's focus on base relations by exploring more column combinations. Some patterns introduce columns or transformations that do not change the original pattern fundamentally. For instance, joining outlier columns with correlated ones may yield weaker but still notable association rules. Unexpected patterns also arise from derived columns, like multiplying an outlier column with increasing datetime values, or generated cluster labels showing strong correlations with certain features.
This observation suggests future work on conditioning MCTS behaviour and interestingness evaluation based on patterns distance.

These results confirm AIDE's expected behaviour, focusing effort on promising search space regions. The generated patterns emerge without specific data transformations and are mostly independent, favouring a shallow search that easily moves across different regions, like the random search. However, experiments show that minor transformations from strong initial patterns retain significant interestingness. While the random policy does not explore in detail these regions, AIDE in UCT configurations consistently explores these areas and uncovers more interesting patterns.

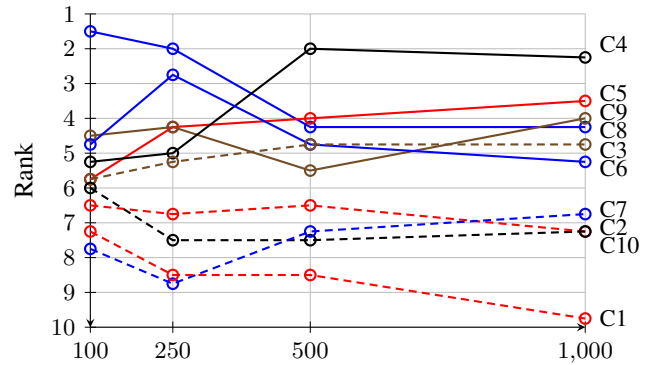
\begin{figure}[t]
\centering
\begin{tikzpicture}
  \begin{axis}[
      width=0.45\textwidth,
      height=5.75cm,
      ymin=1, ymax=10,
      xtick={100,250,500,1000},
      ytick={1,2,3,4,5,6,7,8,9,10},
      ylabel={Rank},
      y dir=reverse,
      grid=major,
      axis x line=bottom,
      axis y line=left,
      clip=false,
      tick style={draw=black},
      label style={font=\normalsize},
      ticklabel style={font=\small},
  ]
    \addplot+[thick, mark=o] coordinates {(100,1.50) (250,2.00) (500,4.25) (1000,4.25)};
    \node[anchor=west, font=\small, xshift=2pt, yshift=-1pt] at (axis cs:1000,4.25) {C8};

    \addplot+[thick, mark=o] coordinates {(100,5.75) (250,4.25) (500,4.00) (1000,3.50)};
    \node[anchor=west, font=\small, xshift=2pt, yshift=3pt] at (axis cs:1000,3.50) {C5};

    \addplot+[thick, mark=o] coordinates {(100,4.50) (250,4.25) (500,5.50) (1000,4.00)};
    \node[anchor=west, font=\small, xshift=2pt, yshift=2.5pt] at (axis cs:1000,4.00) {C9};

    \addplot+[thick, mark=o] coordinates {(100,5.25) (250,5.00) (500,2.00) (1000,2.25)};
    \node[anchor=west, font=\small, xshift=2pt, yshift=5pt] at (axis cs:1000,2.25) {C4};

    \addplot+[thick, mark=o] coordinates {(100,4.75) (250,2.75) (500,4.75) (1000,5.25)};
    \node[anchor=west, font=\small, xshift=2pt, yshift=-2.5pt] at (axis cs:1000,5.25) {C6};

    \addplot+[thick, mark=o] coordinates {(100,6.50) (250,6.75) (500,6.50) (1000,7.25)};
    \node[anchor=west, font=\small, xshift=2pt, yshift=1.5pt] at (axis cs:1000,7.25) {C2};

    \addplot+[thick, mark=o] coordinates {(100,5.75) (250,5.25) (500,4.75) (1000,4.75)};
    \node[anchor=west, font=\small, xshift=2pt, yshift=-2pt] at (axis cs:1000,4.75) {C3};

    \addplot+[thick, mark=o] coordinates {(100,6.00) (250,7.50) (500,7.50) (1000,7.25)};
    \node[anchor=west, font=\small, xshift=2pt, yshift=-5pt] at (axis cs:1000,7.25) {C10};

    \addplot+[thick, mark=o] coordinates {(100,7.75) (250,8.75) (500,7.25) (1000,6.75)};
    \node[anchor=west, font=\small, xshift=2pt, yshift=2pt] at (axis cs:1000,6.75) {C7};

    \addplot+[thick, mark=o] coordinates {(100,7.25) (250,8.50) (500,8.50) (1000,9.75)};
    \node[anchor=west, font=\small, xshift=2pt, yshift=0pt] at (axis cs:1000,9.75) {C1};
  \end{axis}
\end{tikzpicture}
\caption{Average ranking over the 5 synthetic datasets by number of iterations.}
\label{fig:bumpchart_final}
\end{figure}

Figure~\ref{fig:bumpchart_final} summarizes for each step $s$ the average ranking of the configurations across the five datasets in number of interesting patterns found.
No configuration consistently outperforms others, indicating each strategy has strengths and weaknesses based on dataset size and pattern distribution. Promising future work includes developing heuristics to switch configuration parameters based on search evolution. In the results we can observe that C10 underperforms with larger datasets, suggesting a need for hyperparameter tuning of constants $C$ and $D$. C3 generally surpasses C2, highlighting the benefits of the simulation strategy. In action selection, the UCT-based policy (C6) performs better at lower iterations while weighted random (C4) becomes at 500 and 1000 iterations the best performing method, with random action selection (C3) more stable between the two. The fixed number approach for children policy performs well overall, with C8 performing better at fewer iterations and C9 closing the gap at 1000 steps. Among growth rates, 0.25 (C5) is the most stable and effective.

Experiments were run on an 8-core, 2.8 GHz CPU machine with 32GB RAM. As for runtime, the average completion times for 1000 iterations across configurations C2 to C9 are $234.7\pm 46.5$ seconds for $\mathcal{SD}_A$, $1049.4\pm 200.7\,s$ for $\mathcal{SD}_B$, $1742.6\pm252.5\,s$ for $\mathcal{SD}_C$, $3316.4\pm407.1\,s$ for $\mathcal{SD}_D$ and $7827.9\pm977.7\,s$ for $\mathcal{SD}_E$. We aggregated C2 to C9 because there are no significant runtime differences across these configurations, while C1 and C10 consistently exhibit lower runtimes across the datasets. Notably, C7 is an outlier in $\mathcal{SD}_A$ with a runtime $51.4\%$ lower than C1. These differences are closely linked to the number of model actions executed, which is  significantly lower on average for C1 and C10, and C7 in $\mathcal{SD}_A$. This connection is further supported by the number of interesting patterns identified (Figures~\ref{fig:count_plot} and~\ref{fig:bumpchart_final}). Given the high computational cost of the pattern mining actions, it is reasonable to observe that the main factor behind runtime is the number of such actions rather than the the policy itself.

The experiments confirm AIDE can recognize interesting relations within the data (Q1) and that both UCT-based policies and the non-random simulation outperform the random baselines (Q2) Among the different configurations, C4 emerges as the best choice on average (Q3), but different parameters may be more suited depending on the number of steps and the dataset.
 
\section{Related work}\label{sec:related_work}
\textbf{Insights discovery.} The setting and challenges in our paper are also considered in~\cite{top_k_insights}, which highlights limitations of OLAP tools \cite{olap} that rely on user input, proposing an algorithm to automatically extract top-k insights from multi-dimensional data. This approach tackles challenges central to AIDE, such as navigating a vast search space and the computational costs of identifying meaningful insights. Research from \cite{top_k_insights,another_microsoft,quickinsights,another_microsoft2,metainsights} explores subspaces, identifies patterns through aggregation functions, and determines interestingness by highlighting exceptions or trends.
Subspaces are combinations of value-based column filters, therefore, these works do not consider feature generation as we do. Insights are extracted from subspaces via aggregate values, while AIDE uses a broader range of data mining techniques to uncover data relationships.

These studies identify ``exceptional'' and ``outstanding'' values as key indicators of interestingness, aligning with the argument that peculiarity is the only quantifiable dimension in an unsupervised setting. They define measures such as \emph{impact}, which reflects the data proportion within a subspace, and \emph{significance}, assessing exceptionality through p-values.
The subspace exploration methods differ significantly from ours, utilizing custom search techniques based on recursive enumeration and priority queues. They terminate when the time budget is exhausted, similar to our MCTS iterations limit.

\cite{labelled_mcts} proposed an MCTS-based search for data patterns over class labels. Their problem statement differs from ours: the focus is on labelled data to discover data subsets (patterns) that reliably discriminate the given target class. In contrast, we do not assume the presence of a target feature and instead consider a broader family of patterns, in particular using unsupervised pattern mining techniques, such as clustering and association rules, that are not limited to single-feature prediction.

A review of automatic insights tools in~\cite{automated_insights_survey} proposes 12 types of automated insights, including outliers, value/derived value, association, trend, distribution, extreme, cluster, and compound fact. Although insights narration is not our focus, it is crucial for building valuable tools for end users~\cite{storytelling}.

\textbf{Workflow discovery.}
In \cite{ontology_planning, daniel_ml_planning}, the authors focus on the automatic composition of data mining and knowledge discovery workflows through planning, sharing with AIDE the problem of representing and reasoning over actions and preconditions. However, unlike these works, AIDE does not plan for a fixed goal. Similarly, Markov Decision Processes~\cite{mdps} (MDPs) require a fixed goal and precise world model, which is not applicable in our domain. MDPs handle unavailable actions by rendering them ineffective, whereas we use logical preconditions to formally prevent their selection.

\textbf{AutoML.} AutoML~\cite{automl_book,automl_survey2} aims to automatically build the best-performing ML pipeline for a dataset, primarily for supervised tasks. It involves selecting algorithms, tuning hyperparameters, and optimizing model performance based on minimizing prediction error on labelled data~\cite{automl_survey}. Similarly, in unsupervised domains like clustering~\cite{automl_clustering_survey} or anomaly detection~\cite{automl_anomaly}, the goal is to discover a single best-performing model. This is not our goal, which is more open-ended, aiming to extract interesting relations within the data rather than synthesizing a single ML model. For the same reason, the work on MCTS for hyperparameter selection in AutoML~\cite{automl_mcts} is different from ours, as it focuses on achieving optimal model performance, rather than navigating data subsets and models.

\textbf{Feature engineering.}
Feature selection and generation are subproblems we share with AutoML, but different goals lead to different evaluation methods. Our approach prioritizes local evaluation of underlying patterns, guiding the application of models towards the most promising subsets of features. This contrasts with AutoML, where feature selection, including UCT-based methods~\cite{automl_survey,feature_selection_UCT,feature_selection_recursive}, is evaluated based on the generalization error of a classifier trained on the selected features. This approach is often restricted to the initial features, whereas our method explores derived features and row subsets, expanding the scope of feature engineering. Derivation primitives are basic operations commonly used in feature engineering and have been studied within  AutoML~\cite{DBLP:conf/icdm/KatzSS16,van2017automatic}, where the focus remains on classifier prediction performance, and in other domains such as bioinformatics~\cite{feature_selection_bio} and robotics~\cite{manuela}.
    
\section{Conclusion, Limitations and Future Work}\label{sec:conclusion}\label{sec:limitations_future_work}

\textbf{Conclusion.} We introduced AIDE, a novel knowledge discovery method that efficiently explores data transformations and pattern detection models to identify interesting insights. 
AIDE solves the problem of automatically selecting actions to process data and fit data mining models, automating several steps of the knowledge discovery process. The method is easily expansible with new data transformation or data mining techniques without modifying the search algorithm.
Our experiments show that framing the search process as a MCTS problem, combined with the proposed interestingness evaluation of the results, is effective at discovering implicit and potentially useful relations in the data.

\textbf{Limitations.} AIDE operates in a completely unsupervised setting, which restricts its ability to estimate interestingness based solely on statistical properties. This may not fully align with user goals and expectations, and limits the system's capability to reliably estimate state potential in the simulation step. Additionally, AIDE does not utilize column names to infer data semantics, whereas data scientists rely on these semantics to guide decisions on actions and parameter choices.

\textbf{Future work.} Future enhancements to AIDE will focus on integrating domain knowledge and user feedback to improve pattern evaluation. We plan to use Large Language Models (LLMs) to enrich MCTS with common sense knowledge and develop a formal language to express user domain knowledge. Common sense knowledge includes interpreting column names to infer data semantics, which will improve action selection and simulation. User domain knowledge involves goals, expectations, and beliefs, allowing for more tailored pattern evaluation. Additionally, we plan to expand the action set to include more insights discovery techniques, as suggested in~\cite{automated_insights_survey}, and handle multiple tables introducing a join action.

\section*{Disclaimer}
This paper was prepared for informational purposes by the Artificial Intelligence Research group of JPMorgan Chase \& Co. and its affiliates (``JP Morgan'') and is not a product of the Research Department of JP Morgan. JP Morgan makes no representation and warranty whatsoever and disclaims all liability, for the completeness, accuracy or reliability of the information contained herein. This document is not intended as investment research or investment advice, or a recommendation, offer or solicitation for the purchase or sale of any security, financial instrument, financial product or service, or to be used in any way for evaluating the merits of participating in any transaction, and shall not constitute a solicitation under any jurisdiction or to any person, if such solicitation under such jurisdiction or to such person would be unlawful.

\bibliographystyle{IEEEtran}
\bibliography{main}

\end{document}